\documentclass[11pt,a4paper]{article}
\usepackage{booktabs}
\usepackage{hyperref}
\usepackage{cite}
\usepackage{amsmath,amssymb,amsfonts}
\usepackage{algorithmic}
\usepackage{graphicx}
\usepackage{textcomp}
\usepackage{colortbl}
\usepackage{subcaption}
\usepackage[left=2cm,top=1cm,right=2cm]{geometry}
\usepackage[T1]{fontenc}

\usepackage{graphicx,amsmath,bm}

\begin{document}

\title{Encoder Decoder Generative Adversarial Network Model for Stock Market Prediction}


\author{Bahadur Yadav$^{}$\thanks{E-mail: bky0088@gmail.com}, Sanjay Kumar Mohanty $^{}$\thanks{Corresponding author. E-mail: sanjaymath@gmail.com} \\
	\footnotesize{Department of Mathematics, School of Advanced Sciences, Vellore Institute of Technology,}\\  \footnotesize{Vellore 632 014, Tamil Nadu, India}}
\date{ }


\maketitle

\begin{abstract}
Forecasting stock prices remains challenging due to the volatile and non-linear nature of financial markets. Despite the promise of deep learning, issues such as mode collapse, unstable training, and difficulty in capturing temporal and feature-level correlations have limited the applications of GANs in this domain. We propose a GRU-based Encoder-Decoder GAN (EDGAN) model that strikes a balance between expressive power and simplicity. The model introduces key innovations—such as a temporal decoder with residual connections for precise reconstruction, conditioning on static and dynamic covariates for contextual learning, and a windowing mechanism to capture temporal dynamics. Here, the generator uses a dense encoder-decoder framework with residual GRU blocks.  Extensive experiments on diverse stock datasets demonstrate that EDGAN achieves superior forecasting accuracy and training stability, even in volatile markets. It consistently outperforms traditional GAN variants in forecasting accuracy and convergence stability under market conditions.

\end{abstract}


{\bf Keywords:}{Deep Learning, Generative Adversarial Networks, Stock Market Prediction, Time Series Data.}






\section{Introduction }
Forecasting the price of stocks is quite challenging due to the complex interplay of numerous contributing factors, including macroeconomic indicators, political changes, and the financial stability of a given company. To address these complexities, analysts frequently employ two primary forecasting techniques: technical analysis and fundamental analysis. Each technique offers distinct perspectives and tools for understanding and forecasting stock price movement, each with its own theoretical foundations.
In order to find patterns and trends that could predict future market behavior, technical analysis examines past market data, mainly stock prices and trade volumes.  This strategy utilises a range of indicators, including price channels, trend lines, candlestick formations, support and resistance levels, moving averages (MAs), and the Relative Strength Index (RSI).  The basic idea is that historical patterns tend to recur in comparable situations and that market movements reflect the behavior of all investors.  Because of its sensitivity to current events and response to recent market behavior, technical analysis is particularly well-liked by short-term traders.
In contrast, Fundamental analysis takes a longer-term, intrinsic approach by determining a stock's actual value based on its underlying financial facts and general economic conditions.  This entails examining balance sheets, industry trends, corporate earnings, and macroeconomic metrics, including GDP growth, interest rates, and inflation.  Fundamental analysis places more emphasis on qualitative aspects, such as market positioning, corporate strategy, and management effectiveness, than technical methods do.  For long-term investors seeking affordable and reliable stocks, this approach is generally more suitable because it is less sensitive to daily news or market fluctuations. However, both approaches aim to provide insight into potential future stock performance.

Machine learning models have gained popularity in financial forecasting over the past few years due to their ability to identify non-linear relationships within complex datasets. The concept of Generative Adversarial Networks (GANs) has developed significantly since its initial proposal in 2014 \cite{goodfellow2014generative}. The idea of zero-sum non-cooperative games serves as the foundation for the formulation of a GAN as a minimax optimization problem.  It consists of two neural networks that utilise adversarial training: a discriminator and a generator.  While the discriminator learns to differentiate between generated and actual inputs, the generator aims to create synthetic data that closely matches real samples.  Both models advance via this competitive process until the discriminator can no longer accurately distinguish between genuine and fake data, and the generator can successfully replicate the real data distribution. Mathematically, the standard GAN objective is represented as:
\[
\min_G \max_D V(D, G) = \mathbb{E}_{X \sim P_X}[\log D(X)] + \mathbb{E}_{Y \sim P_Y}[\log(1 - D(G(Y)))]
\]
where $X$ represents actual data samples taken from the distribution $P_X$ and $Y$ is a noise vector taken from the previous $P_Y$.  Whereas the discriminator $D(.)$ produces the likelihood that a sample is real, the generator $G(Y)$ converts noise into synthetic data.  Despite its impressive potential, the architecture has some inherent problems, including mode collapse, vanishing gradients, and training instability. Conditional GANs (cGANs) were developed by conditioning on additional inputs \cite{mirza2014conditional}, which retained the instability of the original GAN framework while allowing for control over generating outputs by conditioning on these additional inputs.
To enhance structural adaptability, Deep Convolutional GANs (DCGANs) were presented by \cite{radford2015unsupervised}, which utilize convolutional architectures for improved image generation and stable training. To overcome these limitations, Wasserstein GANs (WGANs) \cite{arjovsky2017wasserstein} were introduced, which addressed these drawbacks by utilising the Wasserstein distance as a more robust loss function. But doing so required weight clipping, which caused optimization problems. By incorporating a gradient penalty into WGAN-GP \cite{gulrajani2017improved}, Lipschitz continuity was developed without clipping. Meanwhile, by penalizing discriminator gradients close to actual data, DRAGAN \cite{kodali2017convergence} reduced instability and improved resilience around data manifolds. By employing an autoencoder-based discriminator, BEGAN \cite{berthelot2017began} introduced equilibrium constraints. By enabling unpaired image-to-image translation through cycle consistency, CycleGAN \cite{zhu2017unpaired} extended the range of applications for GANs. By progressively raising network complexity during training, Progressive GANs \cite{karras2018progressive} enhanced image resolution. In the meantime, BigGAN \cite{brock2018large} was developed, which scaled GANs to produce high-fidelity class-conditional images through extensive training, but at a significant computational cost. By employing style-based architectures \cite{karras2019style} and \cite{karras2020analyzing}, improved generative control over visual characteristics and established new benchmarks in image synthesis.
Despite these developments, the majority of GAN variations have primarily focused on image data. GANs first appeared in the time-series arena in 2019 with TimeGAN \cite{yoon2019time}, which produced realistic sequential data by combining adversarial loss with supervised and reconstructive objectives. Nonetheless, it remained intricate and susceptible to changes in hyperparameters. To represent long-range temporal relationships, TTS-GANs \cite{li2022tts} sought to incorporate transformer-based attention mechanisms into GANs, but they were often computationally intensive. 

To overcome the common drawbacks of GAN-based stock prediction models, this paper presents a unified forecasting framework that addresses these limitations. A customized cost function based on EDGAN is one of the enhancements included in the proposed approach. Further advances include the incorporation of feature matching, conditioning techniques, and windowing procedures.  Based on experimental results, the framework consistently outperforms previous models, such as basic GANs, WGAN-GP, and DRAGAN. However, even with these developments, particularly when compared to more sophisticated hybrid deep learning architectures, there remains an opportunity for improvement. The promising outcomes of this study, however, offer a strong basis for further investigation and advancement in the area of GAN-based time series forecasting. To improve prediction accuracy, this work presents a novel EDGAN-driven model that effectively utilises historical stock data, along with both static and dynamic covariates. The key contributions of this work are summarized as follows:
\begin{enumerate}
    \item In order to forecast stock prices using financial time series, we suggest a new generative framework called EDGAN.  In contrast to basic GANs, EDGAN's generator is supplemented by an encoder module that learns useful latent representations of the input data.  Future values are then reconstructed using these representations, which enhances the generating process.  By addressing the unpredictability produced by noise vectors in basic GANs, this architecture produces outputs that are more precise and controllable.

    \item We incorporate a variety of loss functions into the adversarial setting to enhance training stability and prediction quality.  To guarantee that the discriminator extracts consistent and representative features from both generated and real sequences of data.  This encourages the generator to provide more accurate and coherent forecasts over time by acting as a regularization process.

    \item The proposed EDGAN model is evaluated by extensive tests on real-world stock market datasets.  The empirical findings demonstrate that our method is effective in capturing intricate temporal and market dynamics, as evidenced by its higher forecasting accuracy compared to baseline models.
\end{enumerate}

The rest of the work is organized as follows: Section \ref{2} examines relevant studies on machine learning and deep learning methods for stock price forecasting. The problem formulation is presented in detail in Section \ref{3}. Section \ref{4} presents the components and architecture of the proposed EDGAN-based framework. The experimental setup, results, and comparison with baseline models are provided in Section \ref{5}. The paper is finally concluded, and future work options are discussed in Section \ref{6}.

\section{Related Works }\label{2}
Regression approaches and neural networks are two examples of the models used in machine learning, which has grown in popularity for stock price predictions.  Unlike less frequently updated fundamental measurements, these methods primarily use technical indicators, which provide frequent and structured data.  To manage financial data effectively, time series models have undergone substantial modifications.  The three primary areas of current study in this field are deep learning approaches, traditional machine learning, and classical statistical methods.

\textbf{Traditional Statistical Techniques:} By analyzing financial data as sequential observations, traditional statistical time series models like AR, ARMA, and ARIMA have been used to forecast stock prices.  However, these models frequently have poor forecasting accuracy because market activity is so nonlinear.  To tackle this, hybrid techniques have been investigated, like integrating ARIMA with support vector machines \cite{pai2005hybrid} or augmenting it with neural networks \cite{areekul2009notice}.  Despite these initiatives, machine learning and deep learning alternatives typically outperform compared to traditional statistical methods.

\textbf{Classical Machine Learning Techniques:} For stock market prediction, a number of classical machine learning models have been used, each has different advantages when it comes to managing high-dimensional and non-linear data. The Support Vector Machine(SVM) \cite{fenghua2014stock} was applied for prediction after using Singular Spectrum Analysis (SSA) to identify significant characteristics from the stock data.  In a comparative analysis, SVM, Artificial Neural Network (ANN), Random Forest (RF), and Naive Bayes were utilised to predict the Indian stock market \cite{patel2015predicting}.  According to linear regression \cite{bhuriya2017stock}, it performed better in terms of prediction accuracy than both polynomial and Radial Basis Function (RBF) regressions. To achieve better performance, hybrid and ensemble models have been developed to enhance performance. While  Support Vector - k Nearest Neighbor Clustering (SV-kNNC) improved by substituting self-organizing maps (SOM) for k-means clustering \cite{sinaga2019stock}, 
Later, artificial neural networks were combined with genetic algorithms \cite{ebadati2018efficient}. Fuzzy time series modeling and evolutionary algorithms were applied for improved prediction \cite{zhang2019deep}.
 An adaptive neuro-fuzzy inference system (ANFIS) that combines SVM with artificial bee colony optimization was proposed in order to improve forecasting accuracy \cite{sedighi2019novel}.
Furthermore, a hybrid model for stock index prediction that integrates fractal interpolation and SVM was presented\cite{wang2019novel}. 
 XGBoost and random forests were among the decision tree ensembles used to classify the stock movement directions \cite{basak2019predicting}.
 SVM and k-NN models were fed complex network features for trend prediction \cite{cao2019stock}.

\textbf{Deep Learning Techniques:} The applications of neural networks for stock price prediction problems has grown in popularity as deep learning advances.  The capacity of deep learning to automatically extract pertinent features and identify intricate nonlinear interactions without the need for manual feature engineering is one of its main advantages \cite{chong2017deep}.  To illustrate the efficacy of deep learning in financial modelling, a combined Principal Component Analysis (PCA) with a Deep Neural Network (DNN) was used to improve forecast accuracy \cite{singh2017stock}.
Convolutional Neural Networks (CNNs), renowned for their effectiveness in signal and image processing, have also found applications in the financial industry \cite{cao2019stock1}. In their investigation of CNN-based models and a CNN-SVM hybrid technique for stock index prediction, they demonstrated that both produced highly predictive outcomes.
For time series forecasting, Long Short-Term Memory (LSTM) networks have emerged as a significant deep learning model. While comparing recurrent models, such as LSTM, GRU, and ordinary RNNs, researchers found that LSTM provided better predictive power for Google stock data \cite{di2017recurrent}. 
Similarly, when LSTM was used on S\&P 500 data, researchers found that it performed better than traditional models like deep feedforward networks, logistic regression, and random forests \cite{fischer2018deep}.
Furthermore, additions and enhancements to the LSTM model have been studied.  To further improve the feature selection process, an attention method was introduced \cite{chen2019exploring}.
LSTM and attention layers were merged with investor sentiment to enhance outcomes, highlighting its importance in stock prediction \cite{jin2020stock}. The researchers demonstrated a technique that used wavelet transforms for denoising, stacked autoencoders for dimensionality reduction, and bidirectional LSTM for final prediction \cite{xu2020stacked}. Using a variational autoencoder (VAE) in combination with LSTM to estimate banking stock prices on the Borsa Istanbul market, further studies illustrated the benefits of combining generative and sequential modeling \cite{gunduz2021efficient}. The application of GANs to stock price forecasting has been investigated recently.  In order to improve the conventional GAN loss by adding directional and forecast error components, authors combined GANs with LSTM-based generators and CNN-based discriminators \cite{zhou2018stock}. 
They outperformed standalone LSTM, ANN, SVM, and ARIMA models, as indicated by their results. Similarly, studies have shown better forecasting results than SVR, ANN, and LSTM by using LSTM as the generator and MLP as the discriminator \cite{zhang2019stock}.
Forecasting based on GANs has been substantially enhanced by the addition of sentiment data.  Using sentiment analysis on Apple Inc. financial news, they integrated sentiment scores into a GAN framework \cite{sonkiya2021stock}. Their model outperformed typical GANs, ARIMA, GRU, and LSTM. Further, the study expanded on this concept by utilizing WGAN-GP in conjunction with a CNN discriminator and GRU generator, while additionally utilizing sentiment data \cite{lin2021stock}.
According to their experiments, WGAN-GP outperformed conventional deep learning models in terms of predicting accuracy. Although deep learning models, particularly LSTM, have emerged as the dominating instruments for financial forecasting \cite{bustos2020stock}, GANs are still a relatively new use in this field. 
Although training GANs for time series data presents certain challenges, previous research indicates that they hold a promising future.  Driven by these results, we propose a new GAN-based forecasting model and evaluate its performance against popular models, including standard GANs, DRAGAN, and WGAN-GP.

Encountered difficulties with time series tasks, such as mode collapse and training instability. These problems were solved by WGAN \cite{arjovsky2017wasserstein} utilizing Wasserstein distance, which was further improved by WGAN-GP \cite{gulrajani2017improved} employing gradient penalty for stable training.  Localized gradient regularization around real data was developed by DRAGAN \cite{kodali2017convergence}, providing enhanced robustness in noisy conditions.  By incorporating feature matching into DRAGAN \cite{nejad2024stock}, the realism of the outputs was enhanced, and the model outperformed baseline models, including LSTM, WGAN-GP, and simple GANs.  To increase accuracy, authors utilised CNN and GRU components to integrate sentiment data into a WGAN-GP system \cite{lin2021stock}. To build on these developments, we suggest a new EDGAN architecture for forecasting time series.  Static and dynamic covariates, as well as historical stock prices, are effectively utilised by our approach to enhance predictive accuracy.  This work integrates deep temporal encoding into the adversarial learning process, aiming to enhance GAN-based forecasting.

\section{Problem Formulation}\label{3}
In this section, we focus on the core challenge of long-term forecasting in multivariate stock markets by considering a dataset that contains time series data for $N$ individual stocks. Each series reflects the price evolution of a specific company or financial asset over time. For the $i$-th stock, the historical prices (also called to as the {look-back window}) are denoted by $y^{(i)}_{1:H}$, while the future values to be predicted (also known as the {forecast horizon}) are represented by $y^{(i)}_{H+1:H+F}$. Given access to past observations, the forecasting model aims to infer future price movements within this horizon. This setup illustrates the inherent complexity of stock price prediction, where accurate modeling of historical patterns is critical for capturing long-term trends.

In stock market forecasting, it is common to have access to both static and dynamic covariates prior to making predictions. We denote the dynamic covariates for stock $i$ at time $t$ as $d_t^{(i)} \in \mathbb{R}^k$, where $k$ is the number of such features. These time-varying inputs may include stock-specific indicators, such as trading volume, price volatility, or sector-level performance, as well as global variables, including the day of the week, national holidays, or macroeconomic signals. In contrast, static covariates---denoted by $s^{(i)}$---capture time-invariant characteristics of a stock, including its listing exchange, market capitalization category, or industry classification. For a forecasting model to perform effectively, it must incorporate both types of information, as they offer valuable context that enhances the model’s ability to capture underlying trends and improve predictive accuracy.

The stock price forecasting model  can be thought of as a function that maps the historical stock prices $y^{(i)}_{1:H}$, 
the dynamic covariates $d^{(i)}_{1:H+F}$, and the static covariates $s^{(i)}$ to an accurate prediction of future stock prices. 
Formally, this can be represented as:
\[
f: 
\left\{ y^{(i)}_{1:H} \right\}_{i=1}^{N},\
\left\{ s^{(i)} \right\}_{i=1}^{N},\
\left\{ d^{(i)}_{1:H+F} \right\}_{i=1}^{N}
\longrightarrow 
\left\{ \hat{y}^{(i)}_{H+1:H+F} \right\}_{i=1}^{N}.
\]

To evaluate the performance of the forecasting model, an appropriate error metric is used. When the Mean Squared Error (MSE) is chosen, it measures how closely the predicted future prices align with the actual values. The MSE across all stocks and forecasted time steps is given by:
\begin{align}
& \text{MSE}\Big( 
\left\{ y^{(i)}_{H+1:H+F} \right\}_{i=1}^{N}, 
\left\{ \hat{y}^{(i)}_{H+1:H+F} \right\}_{i=1}^{N}
\Big) \nonumber \\
&= \frac{1}{NF} \sum_{i=1}^{N} 
\left\| y^{(i)}_{H+1:H+F} - \hat{y}^{(i)}_{H+1:H+F} \right\|_2^2.
\end{align}

\begin{figure*}
    \centering
    \includegraphics[width=18.5cm]{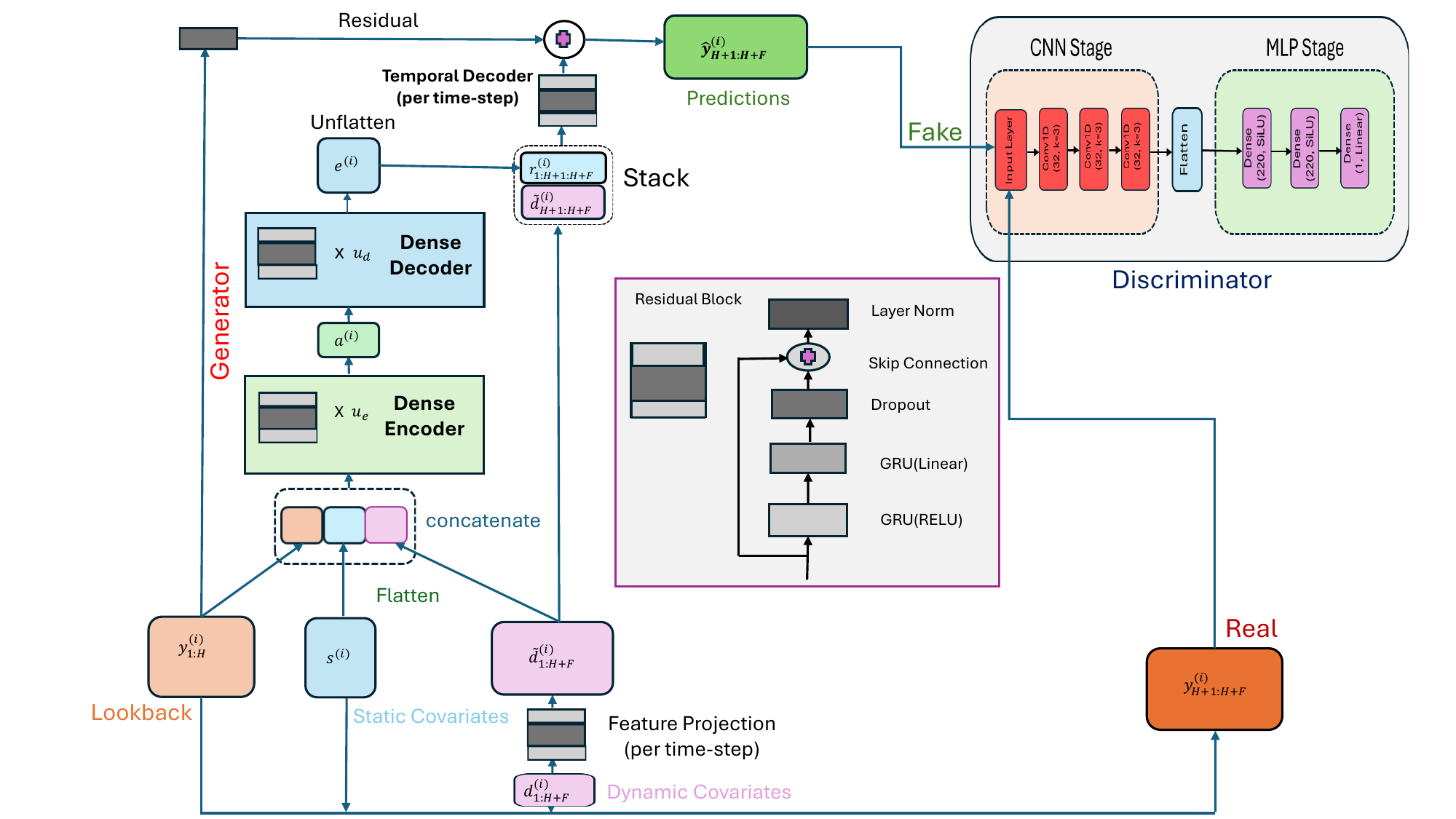}
    \caption{The overall structure of the proposed framework. }
    \label{fig:1}
\end{figure*}

\section{Proposed Model}\label{4}
Recent studies have demonstrated that basic linear models can perform comparably to, or even surpass, Transformer-based models on several long-range forecasting benchmarks.  The incapacity of linear models to represent non-linear relationships between past and future values is a major disadvantage, notwithstanding their unexpected efficacy.  This restriction is particularly significant in complex and erratic fields, such as stock markets, where data-generating procedures are often noisy, highly non-linear, and non-stationary. Furthermore, these models often neglect to include both dynamic and static covariates, as demonstrated by the exclusion of time-varying covariates due to their negative impact on performance.

To address these challenges, we propose a novel architecture called EDGAN, which combines the structured GRU-based encoder-decoder framework with a GAN for reliable and accurate long-term stock market forecasting. In particular, we utilise the encoder-decoder block of EDGAN as the generator, enabling the model to learn intricate covariate-based and temporal patterns in stock prices. With the help of this integration, the system can both predict and model a realistic generative distribution over future price trajectories, contingent on auxiliary features and past market trends.

The generator receives as input the historical stock data \( y^{(i)}_{1:H} \),  static covariate \( s^{(i)} \) and dynamic covariates \( d^{(i)}_{1:H+F} \). These inputs are concatenated and flattened before being passed through a dense encoder to generate a compact latent representation. This representation is then combined with the future dynamic covariates and decoded to produce the predicted future prices $\hat{y}^{(i)}_{H+1:H+F}$. To effectively capture both short-term and long-range dependencies, the generator leverages GRU-based residual blocks enhanced with skip connections, dropout layers, and layer normalization.

To guide the generator in generating realistic sequences, we develop a discriminator that performs the adversarial role of differentiating between real and generated stock price sequences. The discriminator architecture first extracts local temporal features throughout the time horizon using CNN layers.  To capture global sequence-level patterns, these features are then processed by multiple layers of MLPs. Finally, both the predicted future prices (considered as fake) and the actual historical prices (considered as real) are provided as inputs to the discriminator (as illustrated in Figure \ref{fig:1}). In order to achieve the objectives of long-term stock price forecasting, the generator and discriminator are collaboratively optimized in an adversarial way using customized loss functions.

\textbf{Residual Block.}The residual block serves as the basic building block for both the encoder and decoder networks in our proposed EDGAN architecture for stock market forecasting.  A single hidden layer with a ReLU activation makes up each residual block's implementation as a GRU.  By directly connecting the input and output, a linear skip connection encourages gradient flow and helps preserve data from previous levels.  Following the linear transformation in the output layer, we apply dropout to improve generalization and avoid overfitting, which is particularly important in erratic domains like the stock market.  In order to stabilize training dynamics and minimize internal covariate shift, we also use layer normalization at each residual block's output.

The EDGAN model architecture is basically organized into two main components: the generator and the discriminator. The generator is further divided into two distinct phases: an encoding section, followed by a decoding section. The encoding phase features an innovative feature projection step, followed by a dense GRU-based encoder. While decoder phase consists of a dense decoder followed by a unique temporal decoder.
The dense encoder (represented by the light green block with $u_e$ layers) and the dense decoder (represented by the sky blue block with $u_d$ layers) in Figure \ref{fig:1} can be merged into a single block. Since we adjust the hidden layer size in the two blocks independently, we keep them apart for expositional purposes.

\subsection{Generator}
In our framework, the generator is responsible for predicting future stock prices and is organized around two primary parts: a dense encoder and a dense decoder. These parts together comprise an encoder-decoder architecture that transforms historical stock data and related covariates into future predictions, which are measured by the discriminator during adversarial training.

\subsubsection{ Encoding}
The goal of the encoding step is to convert the historical values and covariates of a stock market time series into a dense latent representation that captures auxiliary data and temporal dependencies, forming the basis for producing precise forecasts. In our model, the encoding step consists of two essential components:

\textbf{Feature Projection. }
We apply a residual block to project the dynamic covariates \( d^{(i)}_t \) at each time step (both in the look-back and forecast horizon) into a lower-dimensional space of size \( \tilde{k} \ll k \). This transformation reduces the dimensionality of the temporal features, helping to control model complexity. The transformation is defined as:
\begin{equation}
    \tilde{d}^{(i)}_t = \text{ResidualBlock}\left( d^{(i)}_t \right).
\end{equation}

This projection serves as a learned method of reducing dimensionality. Without it, flattening the entire sequence of dynamic covariates across all $H + F$ time steps would produce a high-dimensional vector of size $(H + F) \cdot k$, which can be computationally demanding and prone to overfitting. Instead, by first mapping each $d^{(i)}_t$ to a lower-dimensional embedding $\tilde{d}^{(i)}_t$, the overall representation is compressed to $(H + F) \cdot \tilde{k}$, enabling more efficient optimization and improved generalization

\textbf{Dense Encoder.}
We begin by stacking and flattening all projected dynamic covariates from both the historical look-back window and the forecast horizon, denoted as $\tilde{d}^{(i)}_{1:H+F}$. This representation is then concatenated with the static covariates $s^{(i)}$ and the historical stock price sequence $y^{(i)}_{1:H}$. The resulting vector is then passed through a dense encoder, composed of several residual blocks, to produce a compact embedding. This encoding process can be expressed as:

\begin{equation}
    a^{(i)} = \text{Encoder}\left( y^{(i)}_{1:H}; \tilde{d}^{(i)}_{1:H + F}; s^{(i)} \right).
\end{equation}

The internal structure of the encoder comprises several residual blocks, each configured with a layer width equal to the designated hidden size. The depth of the encoder, i.e., the total number of residual layers, is determined by a tunable hyperparameter $u_e$, corresponding to encoder layers. The resulting embedding $a^{(i)}$ encodes a unified representation of the stock’s historical behavior, its static covariates (such as sector or company-level information), and relevant time-dependent covariates. This embedding acts as the primary input to the decoding stage.

\subsubsection{Decoding}
In the proposed model, the decoding stage transforms the learned latent representation into forecasts for future time steps of the stock price sequence. This stage consists of two essential components: the dense decoder, which processes the encoded information into an intermediate form, and the temporal decoder, which refines this output to generate time-aware predictions across the forecast horizon.

\textbf{Dense Decoder.} The initial step in the decoding process involves a stack of residual layers, similar in structure and size to those used in the encoder. The decoder takes the latent representation $a^{(i)}$ as input and transforms it into a vector $e^{(i)} \in \mathbb{R}^{F \cdot q}$, where $q$ represents the output dimension per time step. This vector is subsequently reshaped into a matrix $R^{(i)} \in \mathbb{R}^{q \times F}$, with each column $r^{(i)}_t$ corresponding to the model's prediction for the $t$-th time step in the forecast horizon. Formally, the operation is defined as:

\[
e^{(i)} = \text{Decoder}(a^{(i)}) \in \mathbb{R}^{F \cdot q}
\]
\[
R^{(i)} = \text{Reshape}(e^{(i)}) \in \mathbb{R}^{q \times F}
\]

The number of residual layers in the decoder is denoted by $u_d$, which is controlled by the decoder layers hyperparameter.

\textbf{Temporal Decoder.} To produce the final stock price forecasts, we employ a temporal decoder that refines each time-step's output. Specifically, this component is implemented as a residual block with an output dimension of 1. At each forecast time step $t$, it takes the decoded feature vector $r^{(i)}_t$, concatenates it with the projected future covariates $\tilde{x}^{(i)}_{H+t}$, and maps the result to the final predicted value:
\[
\hat{y}^{(i)}_{H+t} = \text{TemporalDecoder}\left( r^{(i)}_t ; \tilde{x}^{(i)}_{H+t} \right), \quad \forall t \in [F].
\]

This operation serves as a direct "shortcut" for future covariates to influence the corresponding forecasted time step, allowing critical external signals—such as macroeconomic announcements or scheduled financial events—to more immediately affect specific predictions. For example, an anticipated interest rate decision could have a sharp effect on asset prices at a known point in the future, which the model can learn more efficiently through this design. The hidden layer size in the temporal decoder is governed by the hyperparameter temporal decoder hidden.

Additionally, we introduce a global residual connection from the input sequence. This is achieved by linearly projecting the historical price sequence $y^{(i)}_{1:H}$ into a vector of the same length as the forecast horizon and adding it to the model’s output $\hat{y}^{(i)}_{H+1:H+F}$. This ensures that simple linear baselines are effectively encompassed as a special case within our architecture.

The generator loss function  is defined as:
\begin{equation}
\mathcal{J}_G = 
- \mathbb{E}_{\mathbf{y}_{\text{fake}}} \left[ \log D(\mathbf{y}_{\text{fake}}) \right].
\end{equation}

\subsection{Discriminator.} The generator is responsible for producing predicted stock prices over the forecast horizon, denoted by \( \hat{y}^{(i)}_{H+1:H+F} \). These predicted sequences are then combined with the historical stock data \( y^{(i)}_{1:H} \), static covariates \( s^{(i)} \), and the future dynamic covariates \( d^{(i)}_{1:H+F} \) that were used in generating the forecast. This concatenated information forms the synthetic (or generated) input that is passed to the discriminator. For comparison, the real input to the discriminator consists of the true future values \( y^{(i)}_{H+1:H+F} \), along with the same historical prices, static attributes, and covariates. This parallel setup enables the discriminator to learn the distinction between authentic and model-generated sequences, thereby guiding the generator toward producing more realistic forecasts.

The discriminator takes both authentic and synthesized sequences as input and is trained to differentiate between the two. These inputs are first passed through a CNN to extract temporal features, followed by an MLP that models higher-level representations. The final output is a scalar probability indicating the likelihood that the input sequence is genuine. Through this adversarial setup, the generator is pushed to create forecasts that are increasingly realistic and consistent with actual market behavior, thereby enhancing the quality and temporal alignment of its predictions.

The discriminator loss function is defined as:
\begin{equation}
\mathcal{J}_D = 
- \mathbb{E}_{\mathbf{y}_{\text{real}}} \left[ \log D(\mathbf{y}_{\text{real}}) \right] 
- \mathbb{E}_{\mathbf{y}_{\text{fake}}} \left[ \log \left( 1 - D(\mathbf{y}_{\text{fake}}) \right) \right].
\end{equation}

\section{Experimental Results }\label{5}
We discuss our main experimental findings on long-term stock market predicting tasks in this section. To forecast future stock prices, a GAN-based framework that combines a temporal decoder and a dense encoder-decoder generator is used for the evaluation.  Yahoo Finance provided data from 9 publicly traded firms with historical stock records from 2010 to 2020 for the model's testing.  Following the extraction of essential features, 14 variables made up the generator's input. They consist of both static and dynamic variables.  Time-invariant characteristics, such as the industry sector (e.g., technology, automotive) and stock exchange (e.g., NASDAQ, NYSE), are encoded by static variables, which facilitate the capture of structural variations at the firm level. In addition to technical indicators like the Relative Strength Index (RSI), Stochastic RSI, Exponential Moving Average (EMA), logarithmic momentum, Moving Average Convergence Divergence (MACD), and Bollinger Bands, the dynamic features include standard stock market indicators such as historical prices like {High}, {Low}, {Open}, {Close}, {Adjusted Close}, and  {Volume}.
The range \([-1, 1]\) was used to normalize all characteristics in order to guarantee uniform input scaling. We used 30\% of the dataset for testing, while 70\% was reserved for training. After the data was segmented using a sliding window of size 3, three consecutive days of 14-dimensional characteristics were represented by input sequences of shape \(3 \times 14\).  The improved generator, which forecasts the stock price for the following day using a dense encoder-decoder design combined with a temporal decoder, was run over these input segments.
The three-day historical window was concatenated with the anticipated price during adversarial training to create the discriminator's fictitious input.  Likewise, the actual input was created by joining the same historical window with the actual next-day price.  More realistic and temporally coherent stock price predictions are then produced by the generator as the discriminator gains the ability to differentiate between generated and real sequences.

\begin{figure*}[htbp]
    \centering
    \begin{minipage}[b]{0.45\textwidth}
        \centering
        \includegraphics[width=1.3\linewidth]{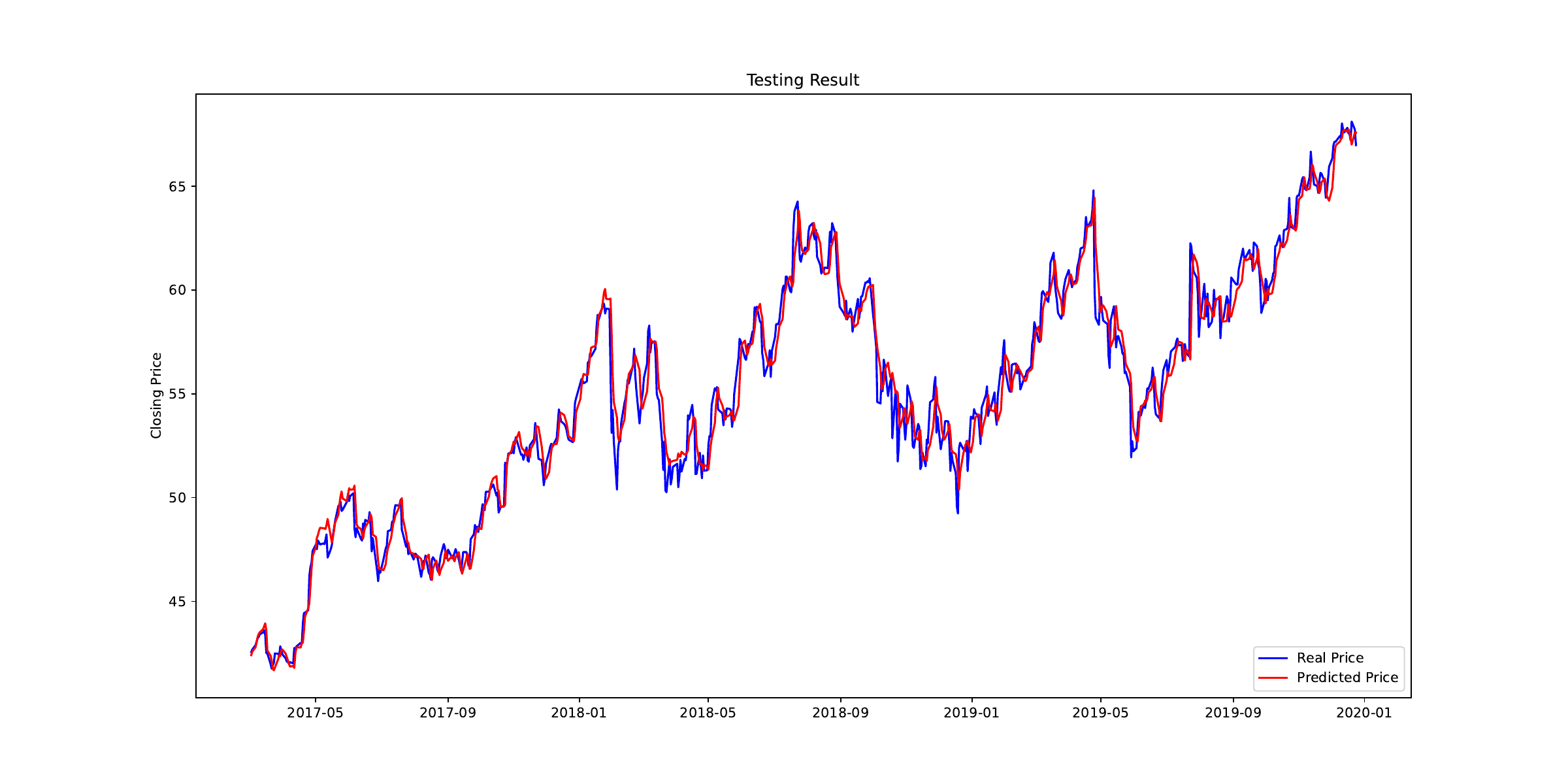}
        \subcaption{Proposed method}
    \end{minipage}
    \hfill
    \begin{minipage}[b]{0.45\textwidth}
        \centering
        \includegraphics[width=1.3\linewidth]{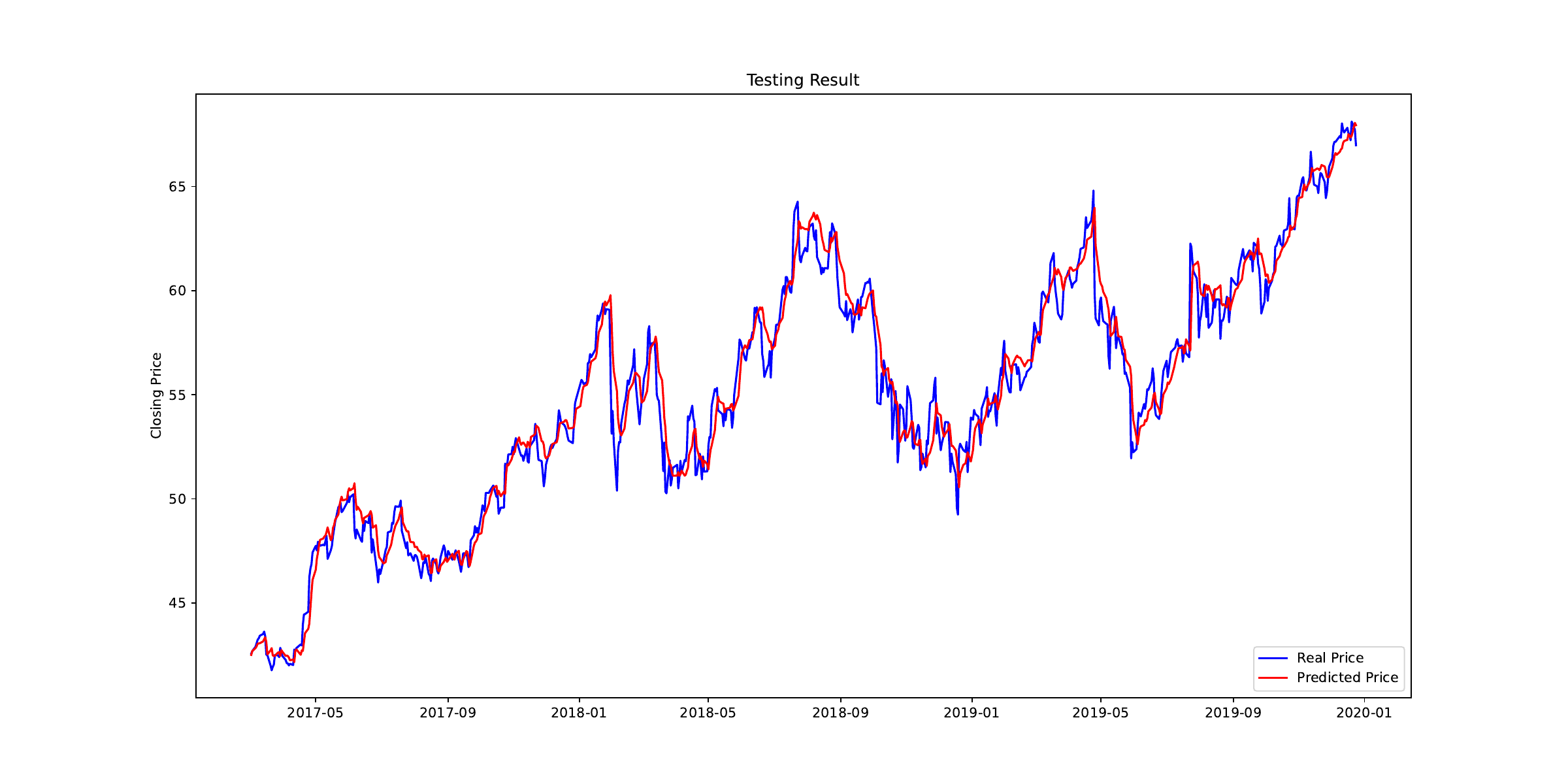}
        \subcaption{DRAGAN}
    \end{minipage}
    
    \vspace{0.5cm}
    
    \begin{minipage}[b]{0.45\textwidth}
        \centering
        \includegraphics[width=1.3\linewidth]{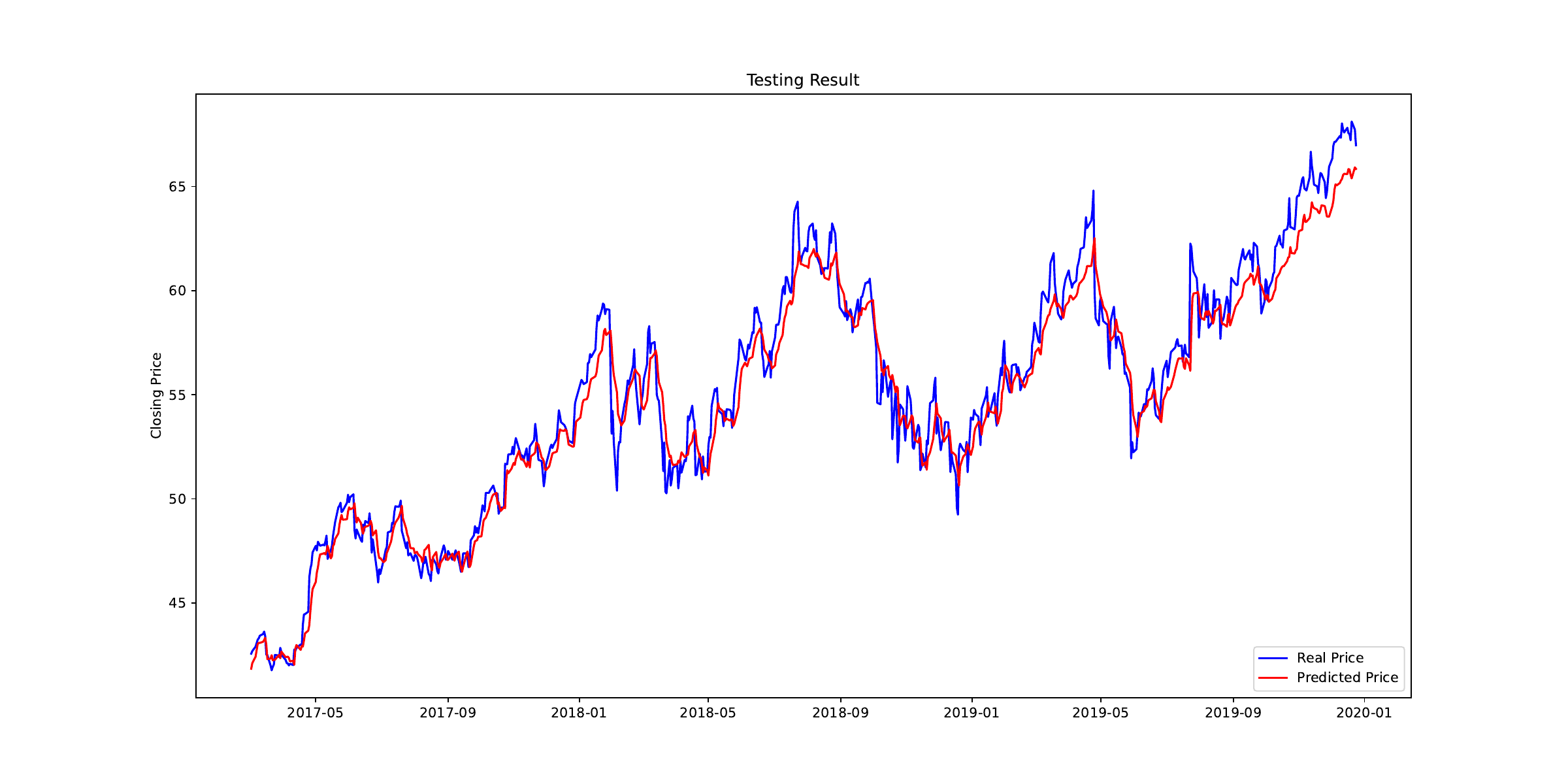}
        \subcaption{WGAN}
    \end{minipage}
    \hfill
    \begin{minipage}[b]{0.45\textwidth}
        \centering
        \includegraphics[width=1.3\linewidth]{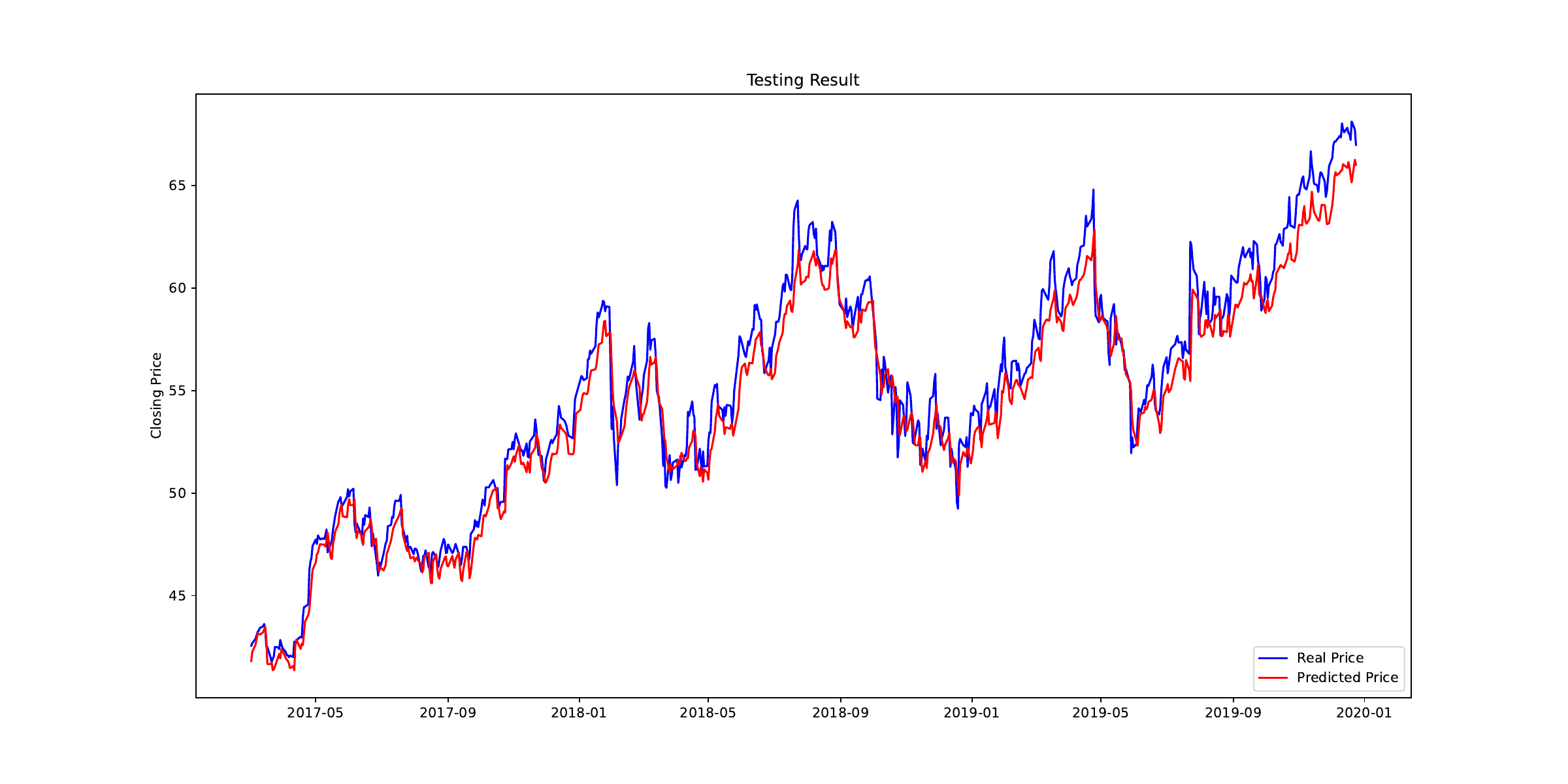}
        \subcaption{Basic GAN}
    \end{minipage}
    
    \caption{Comparison of the real and predicted prices using the test data set (Google stock). (a) Proposed method. (b) DRAGAN. (c) WGAN. (d) Basic GAN.}
    \label{fig:2}
\end{figure*}

\begin{figure*}[htbp]
    \centering
    \begin{minipage}[b]{0.45\textwidth}
        \centering
        \includegraphics[width=1.3\linewidth]{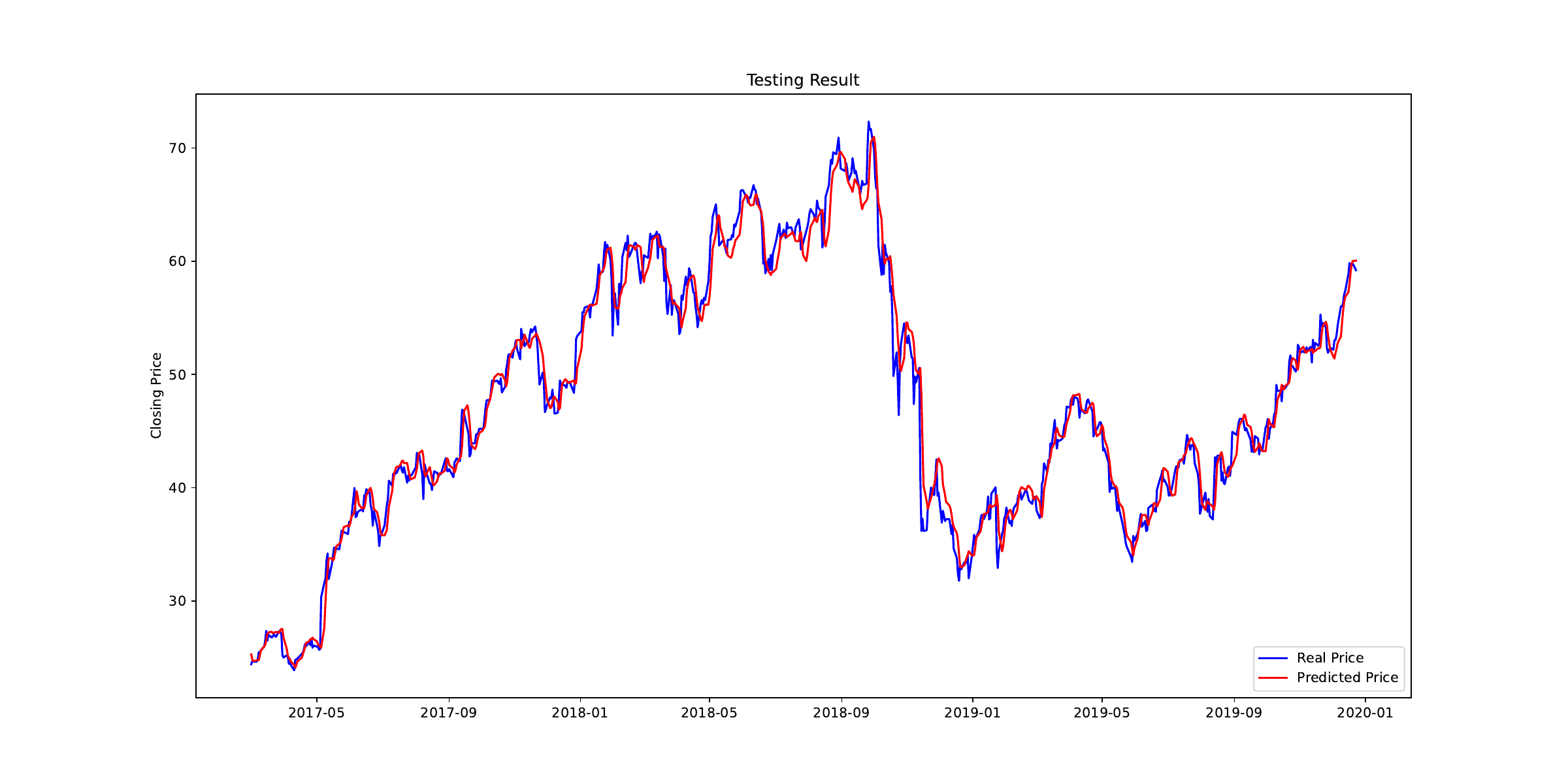}
        \subcaption{Nvidia}
    \end{minipage}
    \hfill
    \begin{minipage}[b]{0.45\textwidth}
        \centering
        \includegraphics[width=1.3\linewidth]{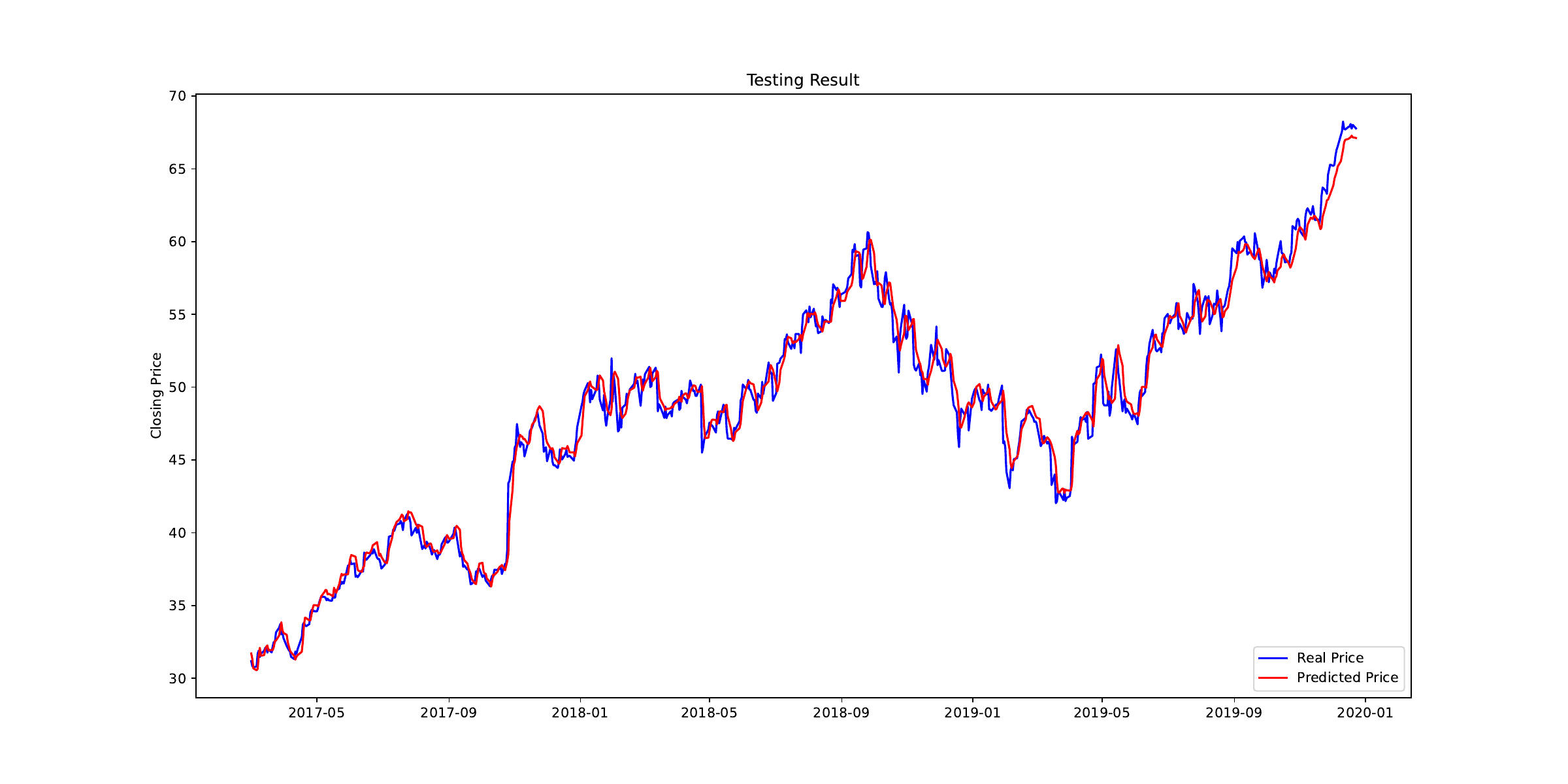}
        \subcaption{Sony}
    \end{minipage}
    
    \vspace{0.5cm}
    
    \begin{minipage}[b]{0.45\textwidth}
        \centering
        \includegraphics[width=1.3\linewidth]{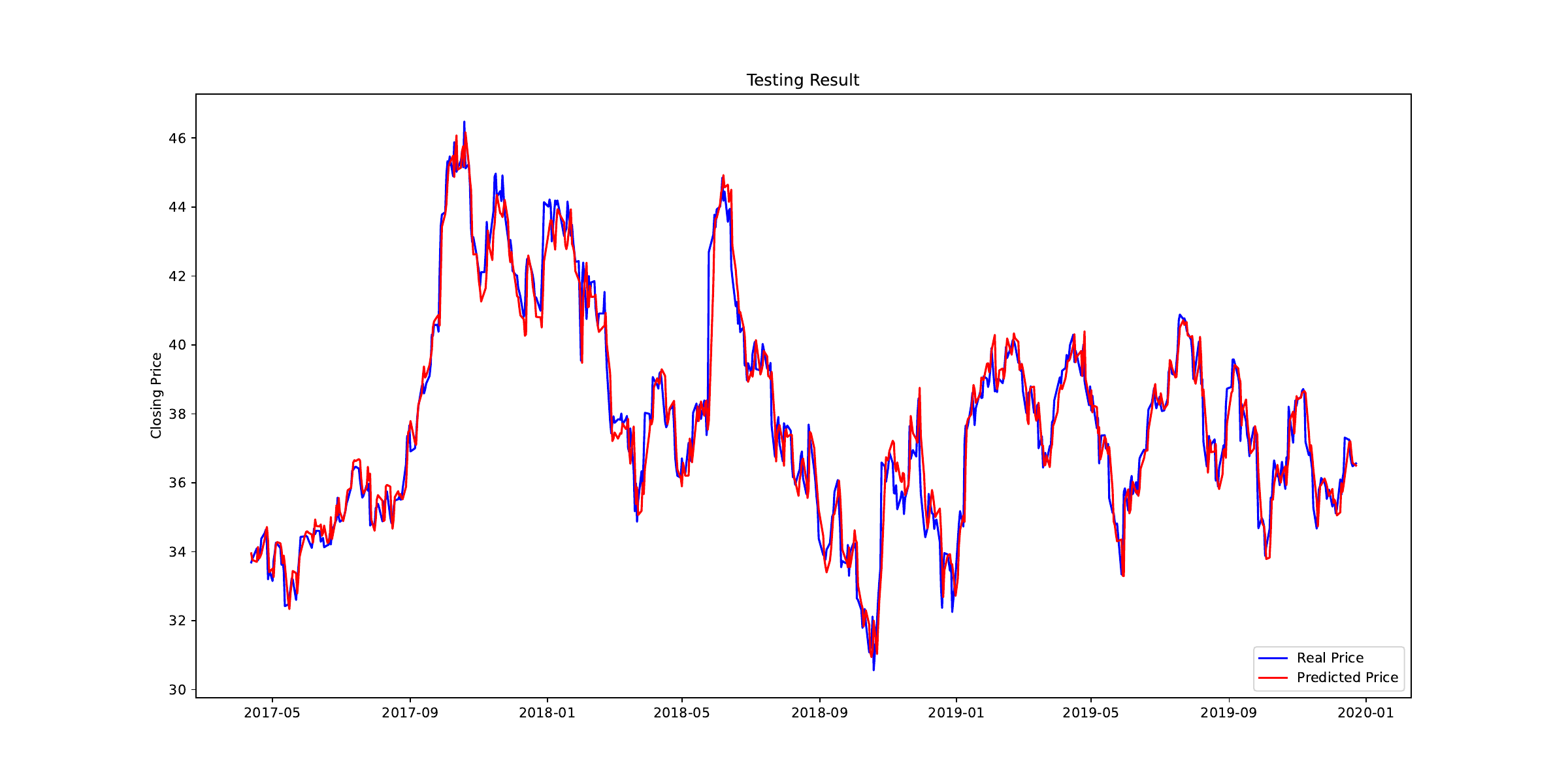}
        \subcaption{General Motors}
    \end{minipage}
    \hfill
    \begin{minipage}[b]{0.45\textwidth}
        \centering
        \includegraphics[width=1.3\linewidth]{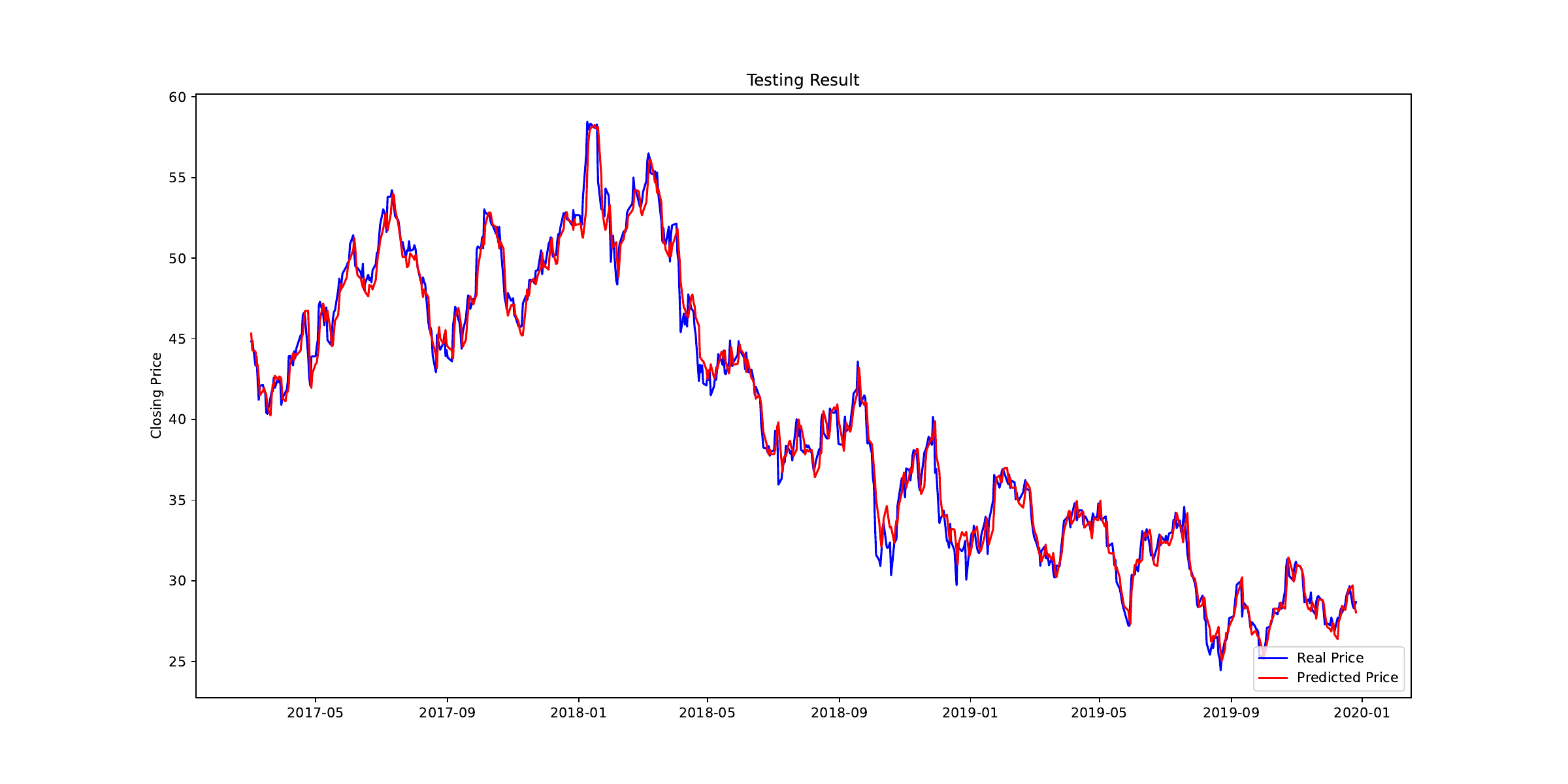}
        \subcaption{AAL}
    \end{minipage}

    \begin{minipage}[b]{0.45\textwidth}
        \centering
        \includegraphics[width=1.3\linewidth]{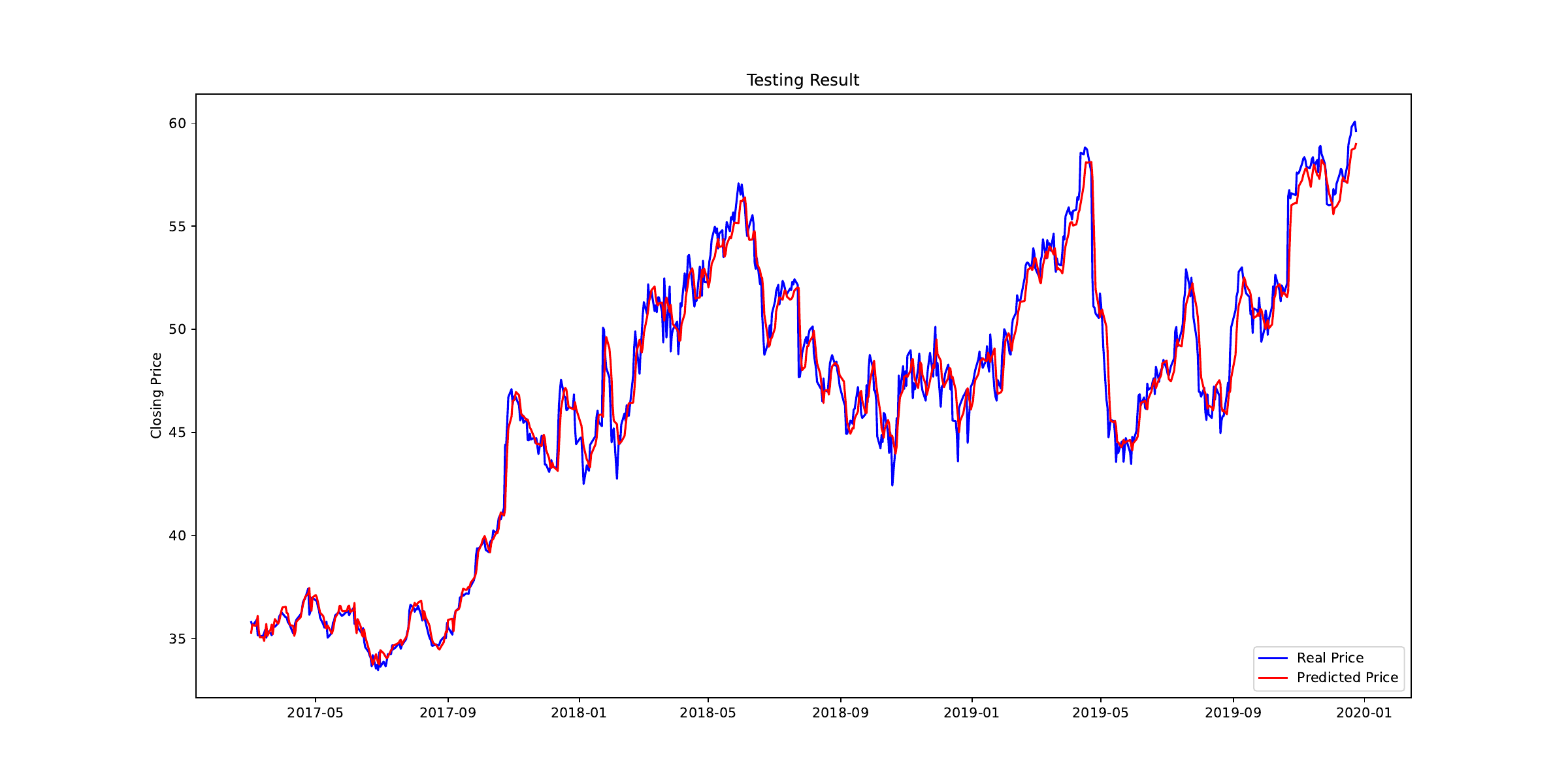}
        \subcaption{Intel}
    \end{minipage}
    \hfill
    \begin{minipage}[b]{0.45\textwidth}
        \centering
        \includegraphics[width=1.3\linewidth]{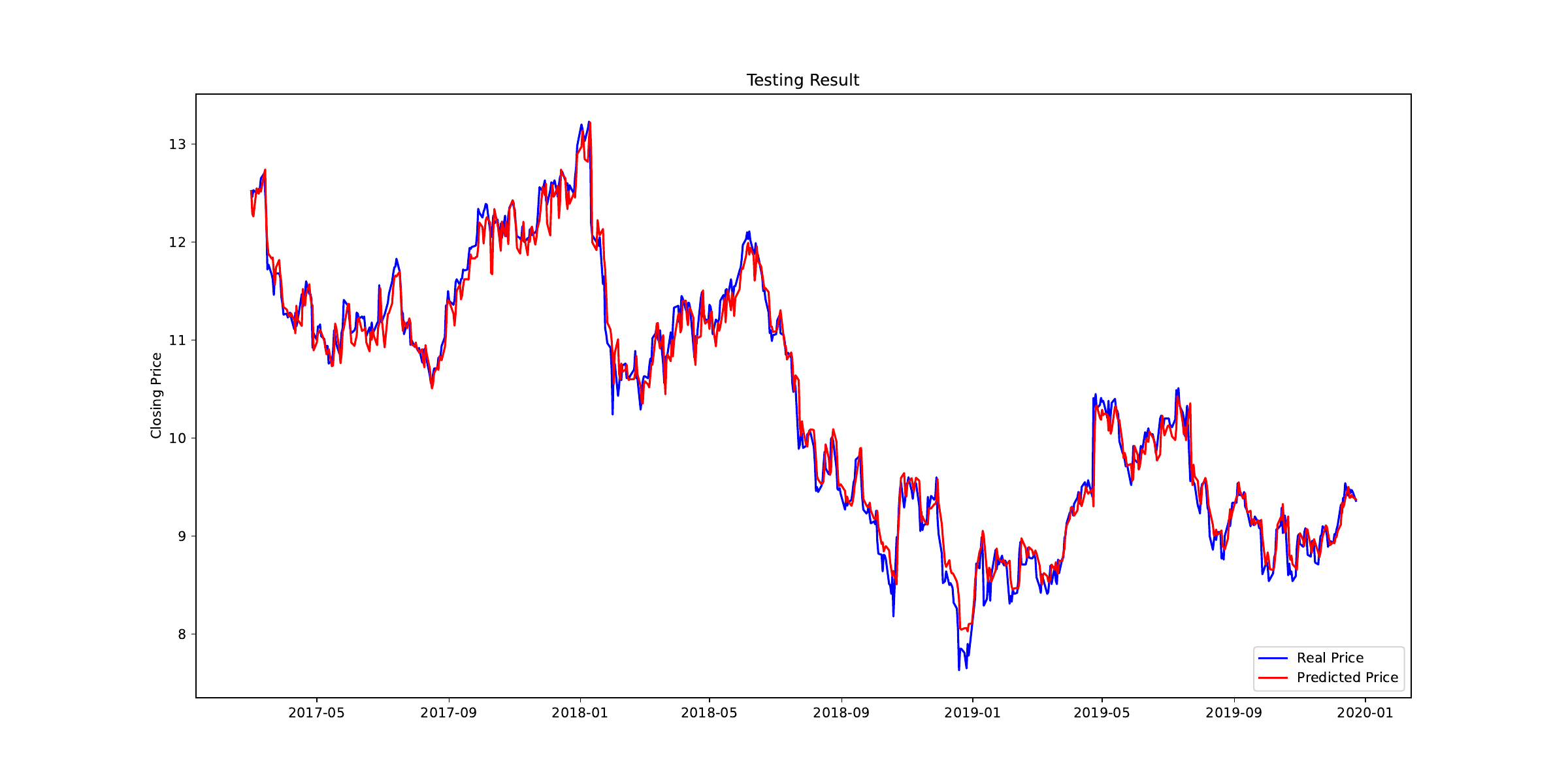}
        \subcaption{Ford}
    \end{minipage}

    \begin{minipage}[b]{0.45\textwidth}
        \centering
        \includegraphics[width=1.3\linewidth]{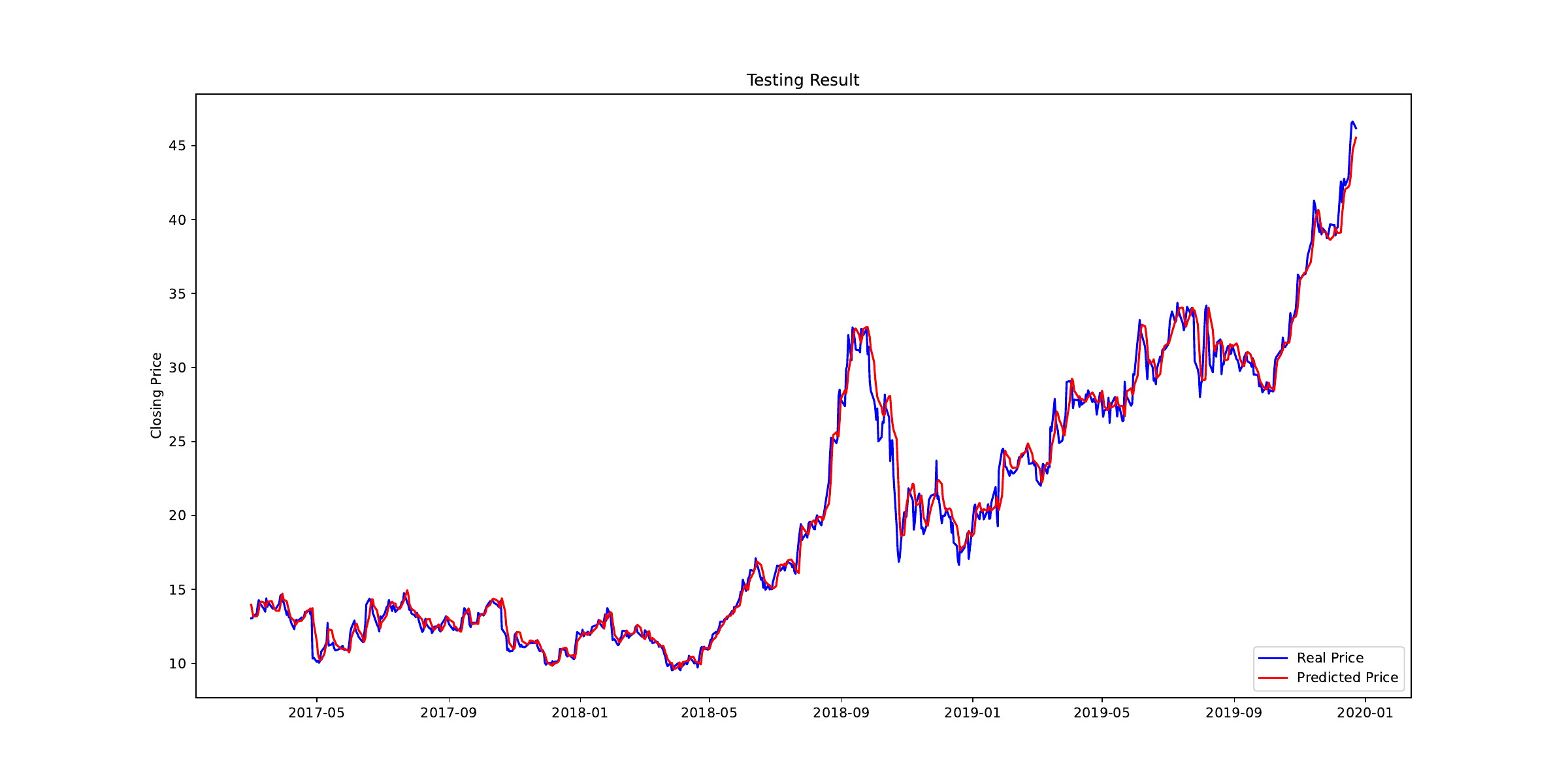}
        \subcaption{AMD}
    \end{minipage}
    \hfill
    \begin{minipage}[b]{0.45\textwidth}
        \centering
        \includegraphics[width=1.3\linewidth]{GOOGL.pdf}
        \subcaption{Google}
    \end{minipage}
    
    \caption{Comparison of the test data set of multiple stocks to compare predicted and real prices. (a) Nvidia. (b) Sony. (c) General Motors. (d) AAL. (e) Intel. (f) Ford. (g) AMD. (h) Google.}
    \label{fig:3}
\end{figure*}

To measure the performance of our proposed model, we examined a comparative analysis against several baselines, including DRAGAN, standard GANs, and WGAN-GP. A summary of the performance metrics for all models is given in Table \ref{tab:1}. As observed, our model consistently outperforms the baselines across key performance metrics. Even though DRAGAN consistently achieves the second-best overall performance, there are some specific situations in which WGAN-GP and simple GANs perform slightly better.  However, the proposed model consistently demonstrates robust and accurate forecasting abilities. Consistency with the underlying performance metrics is demonstrated by the projected stock prices on the test dataset, as shown in Table \ref{tab:1}. We primarily examine the results in terms of RMSE and MAE to facilitate easier comparisons, as Table \ref{tab:1} reveals a high correlation among all three performance metrics.  Furthermore, it is worth noting that in some instances, performance superiority was only demonstrated on the training set. This is undesirable since it indicates poor generalization to data that was not encountered during the testing phase.  This emphasizes the drawbacks of traditional GAN frameworks, which by themselves cannot guarantee reliable or consistent predicting results.  Therefore, while using GANs for stock market prediction tasks, careful consideration of architectural design, training stability, and evaluation technique is crucial. The prediction effectiveness of four models on Google stock prices is shown in Figure \ref{fig:2}.  The proposed method (a) displays higher precision, with predictions that closely match actual prices.  Meanwhile, DRAGAN (b) does well, matching real prices with little variations.  Although it indicates minor variations at peaks, WGAN (c) successfully captures the overall patterns.  The basic GAN (d), on the other hand, exhibits observable prediction errors, particularly when volatility is significant.  The basic GAN yields the least accurate results, whereas the proposed method performs better overall, followed by basic GAN, WGAN, and DRAGAN. These visualisations show that the proposed model can accurately depict underlying stock trends.  Figure \ref{fig:3} illustrates the predicted price trajectories for eight selected stocks, further demonstrating the forecasting capability of the approach.

\begin{table*}[ht]

\centering
\caption{Performance comparison of the proposed method compared to baseline models.}
\label{tab:stock_results}
\begin{tabular}{lllccccc}
\toprule
\textbf{Stock} & \textbf{Phase} & \textbf{Metric} & \textbf{Proposed Method} & \textbf{DRAGAN} & \textbf{WGAN-GP} & \textbf{Basic GANs}  \\
\midrule
Google      & Train & RMSE& 0.46           & 0.86  & 0.55   & 0.52     \\
           &       & MAE &0.31 & 0.68  & 0.40 & 0.38  \\
           &       & R$^2$ & 1.00 & 0.99 &  1.00 &  1.00   \\
           & Test  & RMSE& \textbf{0.95          } & 1.09    &1.21     &1.26      \\
           &       & MAE & \textbf{ 0.68 } & 0.81 & 0.95 & 1.03  \\
           &       & R$^2$ & 0.97 & 0.97 & 0.96&0.95   \\
AAL      & Train & RMSE&  0.85            & 0.98   & 0.94   &0.86      
 \\
           &       & MAE &0.63 & 0.79 & 0.58 & 0.59   \\
           &       & R$^2$ & 1.00 &1.00  & 0.99 &  1.00  \\
           & Test  & RMSE& \textbf{0.96    } & 1.11   &1.12     &1.19     \\
           &       & MAE & {0.73  } & 0.84 & 0.79 &  \textbf{0.50}   \\
           &       & R$^2$ & 0.99 & 0.99 & 0.98 & 0.97  \\

AMD      & Train & RMSE& 0.27             & 0.70 &0.25     & 0.27       \\
           &       & MAE & 0.21&0.60   & 0.17 &0.19   \\
           &       & R$^2$ & 0.99  & 0.96  & 0.99 &0.99  \\
           & Test  & RMSE& \textbf{1.00         } & 1.12   & 1.14  &  0.85    \\
           &       & MAE & \textbf{0.66} & 0.74 &  0.76 & 1.25 \\
           &       & R$^2$ & 0.99  & 0.98 & 0.98 & 0.98 \\
Nvidia      & Train & RMSE& 0.49      & 0.39   &0.48    & 0.35        \\
           &       & MAE &  0.42& 0.28 & 0.33 & 0.26   \\
           &       & R$^2$ &0.99  &0.99  &0.99  & 0.99  \\
           & Test  & RMSE& \textbf{1.64      } &1.68    & 1.76   &1.82    \\
           &       & MAE & \textbf{1.15} & 1.21 & 1.29 & 1.35  \\
           &       & R$^2$ & 0.98 & 0.98 &  0.98&  0.98   \\

Sony      & Train & RMSE& 0.92       & 0.86   & 0.59  & 0.58      \\
           &       & MAE & 0.73& 0.67  &  0.48 & 0.44  \\
           &       & R$^2$ &0.98  & 0.99 & 0.99 &0.98  \\
           & Test  & RMSE& \textbf{1.01      } &1.10   & 1.35   &  1.28     \\
           &       & MAE & \textbf{0.75} &  0.84 & 1.02 &0.99    \\
           &       & R$^2$ & 0.98 & 0.98 &0.98 & 0.97   \\
Intel       & Train & RMSE& 0.52          &0.51    &0.59    &  0.69  \\
           &       & MAE &0.40 & 0.39 & 0.46  & 0.58    \\
           &       & R$^2$ &  0.99  & 0.99 &0.99  &0.98    \\
           & Test  & RMSE& \textbf{ 0.98    } & 1.02   & 1.07   &1.11    \\
           &       & MAE & \textbf{0.71 } & 0.77 &0.84  &0.83    \\
           &       & R$^2$ & 0.98 & 0.98 & 0.98& 0.97  \\

Ford      & Train & RMSE&0.28        & 0.28   &0.29   & 0.30      \\
           &       & MAE &0.20 & 0.21 & 0.22  & 0.23  \\
           &       & R$^2$ &0.98  & 0.98 & 0.98 & 0.98   \\
           & Test  & RMSE& \textbf{0.18      } & 0.20   & 0.20    & 0.21    \\
           &       & MAE & \textbf{0.13} &0.15  & 0.15 & 0.17    \\
           &       & R$^2$ &0.98  &0.98   & 0.98&0.97    \\
General Motors     & Train & RMSE& 0.64       & 0.62   &0.64   & 0.65    \\
           &       & MAE &0.49 & 0.48 & 0.50  & 0.50  \\
           &       & R$^2$ & 0.98  & 0.99 & 0.99 &  0.98   \\
           & Test  & RMSE& \textbf{0.68    } & 0.71   & 0.75   & 0.72     \\
           &       & MAE & \textbf{0.49 } & 0.51 & 0.56  & 0.53  \\
           &       & R$^2$ & 0.95 &  0.95 & 0.94& 0.95  \\
           Apple      & Train & RMSE&  0.41 & 0.84 & 0.43 & 0.41  \\
           &       & MAE &  0.30 & 0.67 & 0.32 & 0.30  \\
           &       & R$^2$& 1.00& 0.99 & 1.00 & 1.00  \\
           & Test  & RMSE& \textbf{ 0.87} & 0.92 & 0.95 & 0.98  \\
           &       & MAE & \textbf{0.63} & 0.68 & 0.69 & 0.72  \\
           &       & R$^2$& 0.99 & 0.98 & 0.99 & 0.99  \\

\bottomrule
\end{tabular}

\label{tab:1}
\end{table*}

\begin{figure*}
    \centering
    \includegraphics[width=14cm]{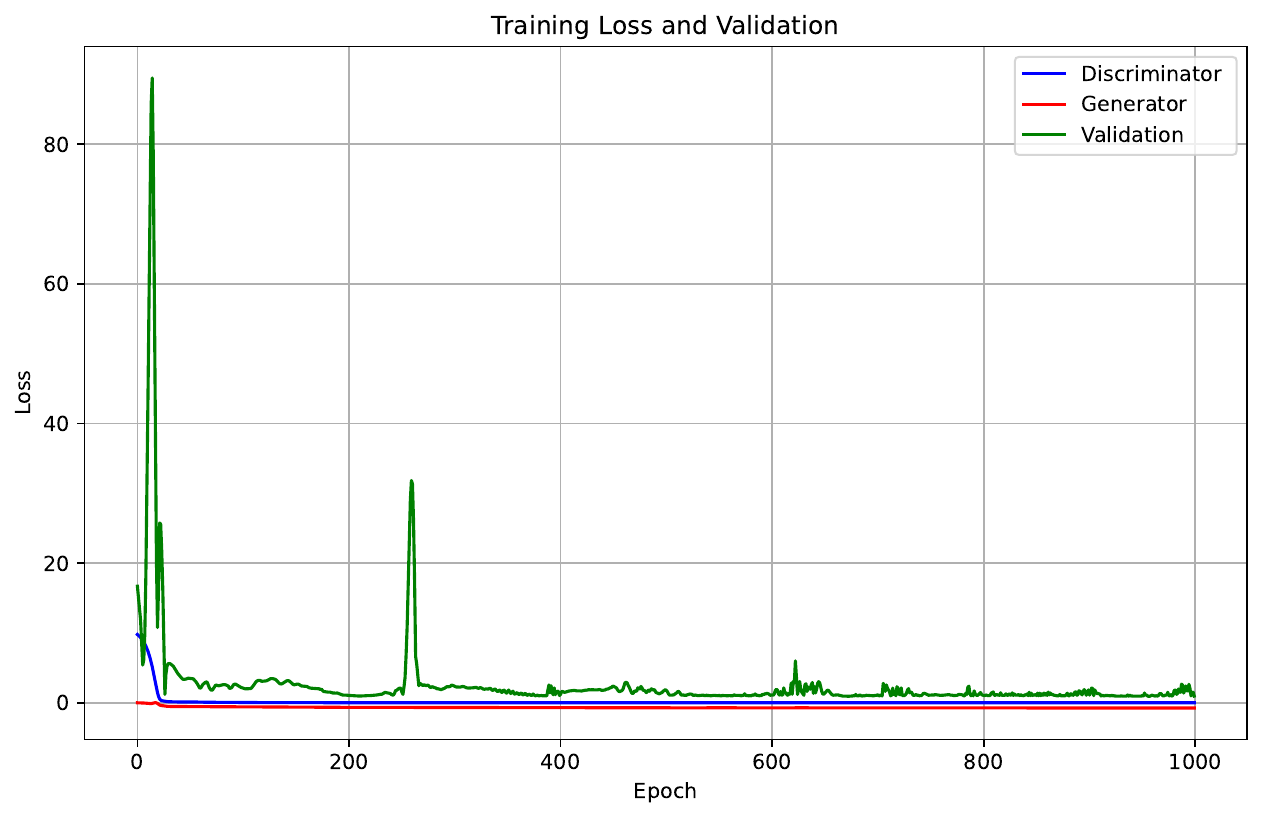}
    \caption{Convergence behavior of the proposed model. }
    \label{fig:4}
\end{figure*}

\textbf{Behavior of the convergence:}
We evaluated the convergence of the discriminator and generator during the adversarial training phase, as illustrated in Figure \ref{fig:4}. Plotting of the validation loss (green), discriminator loss (blue), and generator loss (red) during 1000 training epochs is shown. The validation loss exhibits a notable increase in the initial stages of training, indicating instability as the generator attempts to deceive the discriminator and discover the underlying data distribution. The first drop in the discriminator's loss indicates that it is rapidly adapting and becoming stronger. As training progresses, both the discriminator and generator level off and reach an almost stable adversarial equilibrium. As the generator gradually learns to generate more realistic outputs, the validation loss exhibits a significant decline following the initial spike, eventually flattening out. This pattern demonstrates the generator's gradual improvement and increased capacity to capture the temporal framework of the data. Even though there are still minor variations in the later epochs, particularly in the region of epochs 250–300, the losses remain minimal and exhibit no divergent patterns, indicating convergence.  After stopping training at 1000 epochs, we noticed that the generator continued to generate useful outputs without showing any discernible improvement in validation error.

\section{Conclusion}\label{6}
This work presents a novel encoder-decoder-based generative adversarial architecture for precise stock price forecasting.  The model overcomes the drawbacks of standard GANs, which rely on random noise, by successfully learning latent representations from structured financial time series data through the integration of an encoder into the generator.  The generator's output is regularized and improved by the integration of loss functions, which guarantees the discriminator catches consistent temporal patterns.  Numerous tests on real-world datasets demonstrate that EDGAN outperforms well-known baselines, including standard GANs, WGAN-GP, and DRAGAN, achieving higher accuracy in simulating complex market dynamics.  According to these encouraging findings, EDGAN offers a reliable and efficient approach to developing GAN-based time series forecasting models for use in financial applications.


\section*{Conflict of Interest}
The authors of this article have disclosed that they have no potential conflicts of interest pertaining to the research, writing, or publication of the work.

\section*{Acknowledgement}
S. K. Mohanty acknowledges partial support from the Department of Science and Technology, Government of India, under grant number SR/FST/MS-II/2023/139-VIT Vellore.




\end{document}